\documentclass{article} 
\usepackage{iclr2027_conference,times}
\usepackage{multirow}
\usepackage[table]{xcolor}
\usepackage{algorithm}
\usepackage{algpseudocode}
\usepackage{booktabs}
\usepackage{graphicx}
\usepackage{afterpage}
\usepackage{subcaption}
\usepackage{wrapfig}
\usepackage{enumitem}

\usepackage{xcolor}
\definecolor{promptSlot}{HTML}{ED7D31}
\definecolor{promptThink}{HTML}{0070C0}
\definecolor{promptAction}{HTML}{7030A0}

\usepackage[most]{tcolorbox}
\newtcolorbox{casestep}[1]{
  title={#1},
  colback=white,
  colframe=black!70,
  colbacktitle=black!70,
  coltitle=white,
  fonttitle=\bfseries,
  boxrule=0.6pt,
  arc=2pt
}
\usepackage{float}

\usepackage{booktabs}
\usepackage{multirow}
\usepackage{graphicx}
\usepackage{amsmath}
\usepackage{arydshln}
\usepackage[dvipsnames,table]{xcolor}

\newcommand{\pmstd}[1]{_{\scriptscriptstyle\pm #1}}

\usepackage{amsmath,amsfonts,bm}

\def\eqref#1{equation~\ref{#1}}

\def\1{\bm{1}}

\DeclareMathAlphabet{\mathsfit}{\encodingdefault}{\sfdefault}{m}{sl}
\SetMathAlphabet{\mathsfit}{bold}{\encodingdefault}{\sfdefault}{bx}{n}

\def\sC{{\mathbb{C}}}

\def\sN{{\mathbb{N}}}

\newcommand{\E}{\mathbb{E}}

\usepackage{hyperref}
\usepackage{url}

\title{Depth Before Breadth: Reverse-to-Forward Curriculum with advantage decomposition for LLM Agent}

\author{%
\normalfont
\begin{minipage}{\dimexpr\textwidth-2\tabcolsep\relax}
\centering
\begin{tabular}{@{}c@{\hspace{1.65em}}c@{\hspace{1.65em}}c@{\hspace{1.65em}}c@{}}
Siyuan Du\textsuperscript{1} &
Yuxin Liu\textsuperscript{2} &
Meng Tang\textsuperscript{3} &
Haoran Li\textsuperscript{1} \\[0.75em]
\multicolumn{4}{c}{%
Yechao Bai\textsuperscript{1}\hspace{1.65em}%
Feiyang Pan\hspace{1.65em}%
Jia He} \\[1.15em]
\end{tabular}\\[-0.2em]
{\small
\textsuperscript{1}\,Nanjing University\hspace{1.5em}%
\textsuperscript{2}\,Peking University\hspace{1.5em}%
\textsuperscript{3}\,University of Wales, Aberystwyth}%
\end{minipage}%
}

\iclrfinalcopy 
\begin{document}
\raggedbottom

\maketitle

\begin{abstract}
Reinforcement learning (RL) has become a common approach for training large language model (LLM) agents on long-horizon tasks. However, learning from sparse terminal rewards makes credit assignment and exploration difficult. We approach these challenges through a \emph{policy-induced capability region}, the set of states from which the current policy can reliably complete a task. Near its boundary, an agent may need only a few actions to reach a more solvable state, even when the goal remains distant. This reduces long-horizon learning to two coupled subproblems of reaching that state and completing the task from there. We address them with paired advantages and a progressive reverse-to-forward curriculum. The advantages provide separate signals for reaching more solvable states and completing the task from them, while the curriculum expands the capability region backward along successful reference trajectories and then into off-reference states through on-policy exploration. On ALFWorld and WebShop, our method outperforms competitive agentic RL methods at both model scales, achieving success rates of $90.1\%/96.1\%$ and $80.2\%/82.6\%$ with Qwen2.5-1.5B/7B-Instruct, respectively.
\end{abstract}

\section{Introduction}
\label{sec:intro}
Large language models (LLMs)~\citep{achiam2023,grattafiori2024,deepseek2024v3} are increasingly deployed as agents that interact with external environments on behalf of users~\citep{yao2023,wang2023voyager}.
Moving beyond traditional single-turn generation, these agents operate through a stateful, sequential decision process where each action continuously adapts to evolving observations. Reinforcement learning (RL) has emerged as the dominant paradigm for training such interactive agents\citep{zhou2024archer,qi2025,wang2025,xi2025}, with group-based policy gradient methods, particularly Group Relative Policy Optimization (GRPO)\citep{shao2024}, fast becoming the prevailing standard~\citep{feng2025,zheng2025gspo}. However, scaling these algorithms to long-horizon agent tasks remains bottlenecked by reward sparsity~\citep{andrychowicz2017her,jin2025,yu2026dapo}. All failed rollouts produce no learning signal at any step, and even successful ones deliver only a single terminal reward that has to be attributed across the entire trajectory. This sparse feedback makes credit assignment particularly challenging as LLM agents navigate vast action spaces and diverse tasks through partially observable environments with textual observations.

To address reward sparsity, a standard strategy is to reshape the training distribution through curriculum learning, which exposes the policy to simpler tasks before introducing progressively harder ones~\citep{bengio2009,narvekar2020}. Reverse curriculum strategies operationalize this principle by initializing the agent near the goal and incrementally pushing the starting state farther back~\citep{florensa2017}. Recent LLM agent work adopts a reference-anchored variant~\citep{zhang2026}. Instead of sampling start states by random walks, the start regresses backward along a successful reference trajectory, so the policy first masters local execution near the goal and is only later asked to complete the task from the original start~\citep{resnick2018,tao2024rfcl}. 

However, reverse curriculum leaves a structural credit assignment problem unresolved. Although initializing closer to the goal allows terminal feedback to be attributed more precisely to individual actions, moving the starting point farther back stretches the trajectory, forcing the same coarse feedback to be distributed over a longer sequence. Consequently, isolating early actions that genuinely drove the outcome remains difficult~\citep{arjona2019rudder,pignatelli2023}. A separate limitation is that reverse curriculum starts are anchored to reference trajectories, limiting exposure to the off-reference states encountered during the policy's own exploration~\citep{ross2011}.

\begin{wrapfigure}{r}{0.35\linewidth}
\centering
\includegraphics[width=\linewidth]{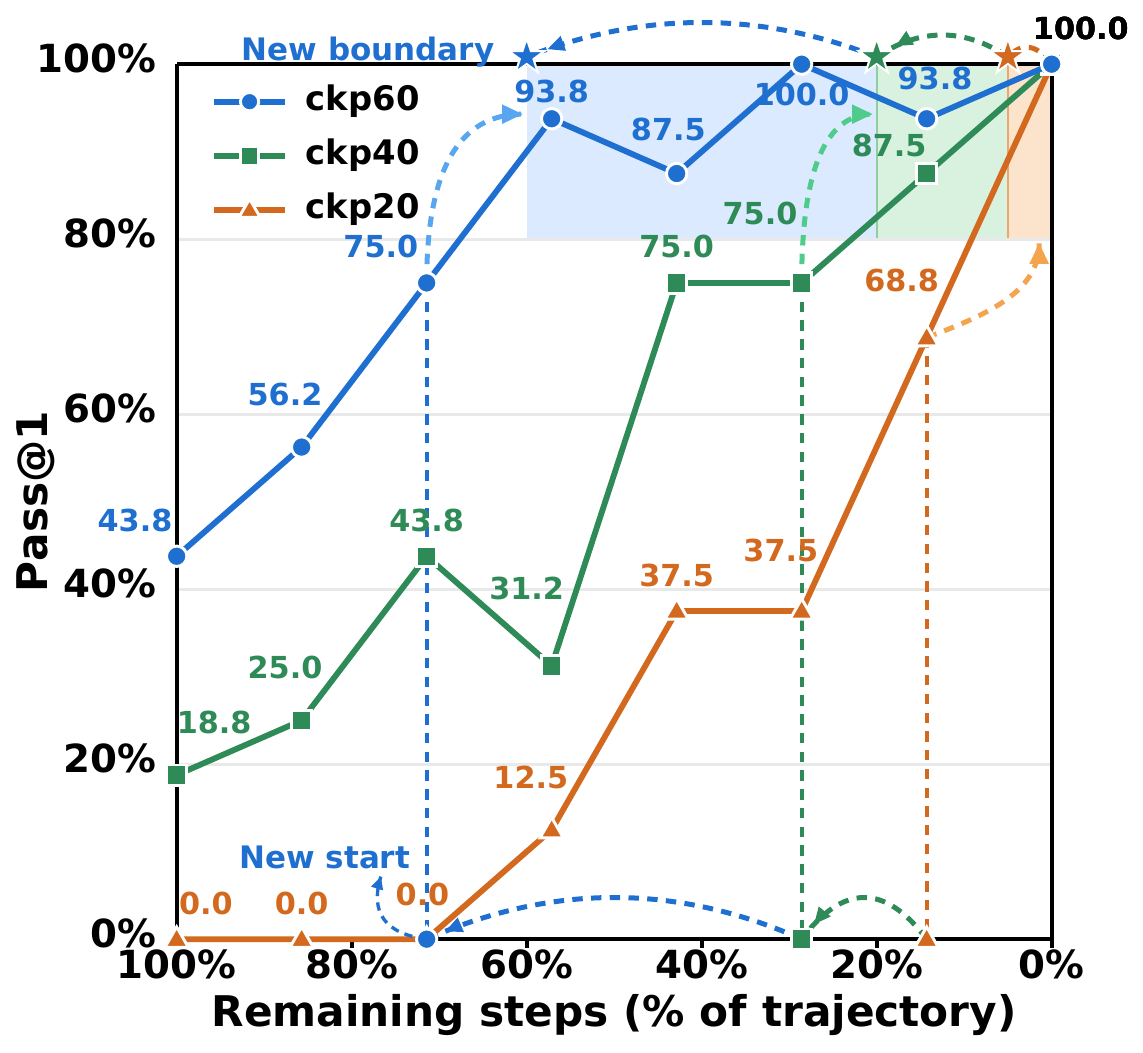}
\caption{The three curves show Pass@1 at checkpoints 20, 40, 60 in reverse curriculum training , as the training start moves backward along a successful reference trajectory from near the goal toward the original task start. Defining the capability region on this trajectory as starting positions with Pass@1 above $80\%$, we observe that this region also expands toward the original start. By ckp60, its new boundary reaches approximately $60\%$ remaining steps.}
\label{fig:decay}
\end{wrapfigure}

Our key insight is that moving the training start backward also expands the policy's \emph{capability region} toward the original start. As Figure~\ref{fig:decay} shows, later checkpoints in reverse curriculum achieve high success from states progressively farther from the goal along the reference trajectory. A newly introduced starting state near this boundary can still be far from the goal, yet only a few actions away from a state within the policy's existing capability region. We exploit this structure by splitting credit assignment into two subproblems: \textbf{(i)} \emph{how to enter} the capability region from a state just outside it, a short-horizon problem; and \textbf{(ii)} \emph{how to complete} the task from the state reached inside it, a longer-horizon problem that earlier training has already partially addressed. The same decomposition can then be applied to states near the capability boundary encountered in the policy's own rollouts, extending training beyond the reference trajectory.

\afterpage{%
\begin{figure}[t]
\centering
\includegraphics[width=0.75\linewidth]{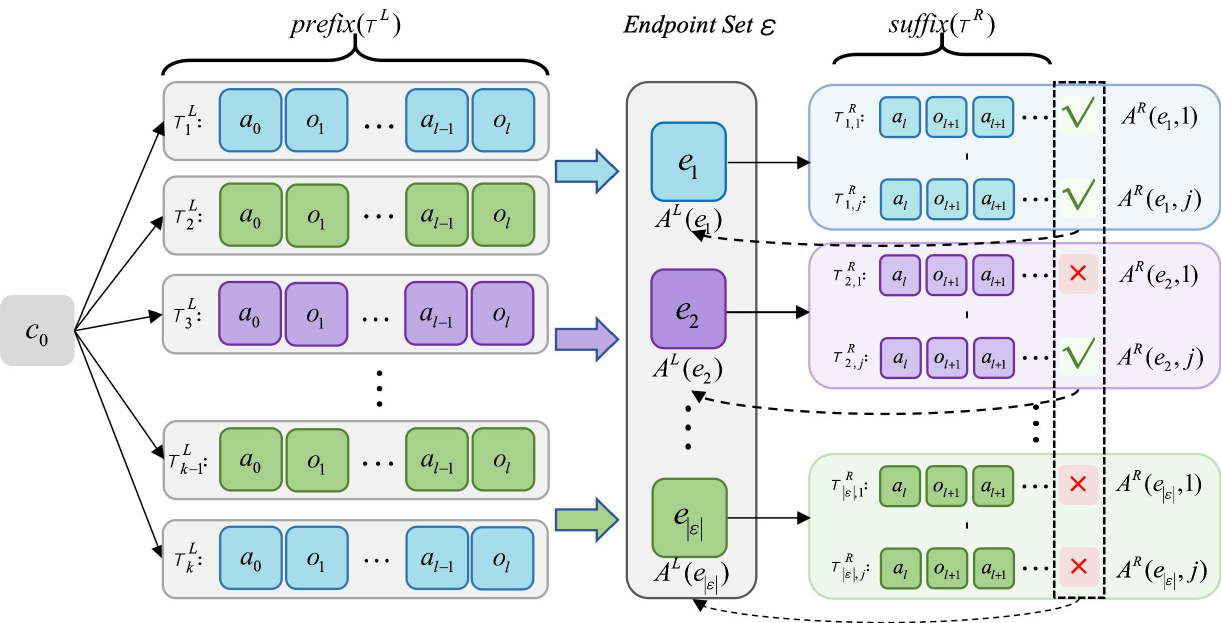}
\caption{\textbf{Overview.} Starting from a context $c_0$, the policy samples $n$ prefixes $\tau_k^{\mathrm{L}}$ and groups those that reach the same endpoint context $e$, yielding the set of distinct endpoints $\mathcal{E}$. From each $e\in\mathcal{E}$, it samples $N_{\mathrm{R}}$ suffixes $\tau_{e,j}^{\mathrm{R}}$ and observes their terminal rewards. The left advantage compares estimated completion rates across endpoints to credit prefixes for where they lead, while the right advantage compares suffix outcomes from the same endpoint to credit the actions taken after reaching it.}
\label{fig:overview}
\end{figure}
}

We operationalize this decomposition through a Paired Advantage (PA) update. For each training group, we sample prefixes from a starting state near the capability boundary, group them by their endpoints, and sample multiple suffixes from each distinct endpoint. The left advantage compares estimated completion rates across prefix endpoints, crediting prefixes that reach states the policy can solve more reliably and thereby guiding it into the capability region. The right advantage compares suffix outcomes from the same endpoint to identify which continuations lead to task completion. The suffix outcomes thus provide both the endpoint estimates used for prefix credit and the comparisons used for suffix credit. We apply this update in two stages. The reverse stage establishes the capability region along successful reference trajectories, moving training starts backward from near the goal until the original starts become reliably solvable for most tasks. The forward stage then broadens the region beyond those trajectories by training on states reached through prefixes sampled by the policy from the original starts. Our contributions are threefold:
\begin{itemize}[
  leftmargin=1.5em,
  labelsep=0.4em,
  itemsep=1pt,
  topsep=2pt,
  parsep=0pt,
  partopsep=0pt
]
\item We introduce a policy-induced capability region view of reverse curriculum training. It reveals a short horizon problem of entering the region within a long horizon task.
\item We propose a paired advantage method with a reverse-to-forward curriculum. Using terminal rewards alone, it compares completion rates across prefix endpoints to guide entry into the capability region and compares suffix outcomes from the same endpoint to improve task completion.
\item Comprehensive experiments on ALFWorld\citep{shridhar2021} and WebShop \citep{yao2022} show that our method outperforms strong baselines and achieves a larger capability region under matched environment interaction budgets.
\end{itemize}

\section{Preliminaries}
\label{sec:prelim}
\paragraph{Problem setup.}
We consider a distribution of tasks $x \sim \mathcal{D}$. For each task $x$, an LLM agent interacts with an environment over multiple steps. At step $t$, the agent receives observation $o_t$ and emits a textual action $a_t$; the next observation $o_{t+1}$ then arrives in response. A trajectory $\tau = \{(o_t, a_t)\}_{t=0}^{T-1}$ together with its terminal reward $R(\tau) \in \{0, 1\}$ captures one full interaction, where $R(\tau)=1$ if the task is completed successfully and $0$ otherwise. The agent’s behavior is governed by an LLM policy $\pi_\theta$ that conditions the interaction. We define a \emph{decision context} $c_t = (\, o_{\leq t},\, a_{< t}, x)$ at step $t$, yielding an action according to $\pi_\theta(a_t\mid c_t)$.
The learning objective is:
\begin{equation}
J(\theta)
=
\E_{x \sim \mathcal{D},\,
\tau \sim \pi_\theta(\cdot \mid x)}
\left[R(\tau)\right].
\label{eq:objective}
\end{equation}
\paragraph{Group-based policy optimization.}
Group-based methods estimate relative advantages by comparing several rollouts of the same task. Given a task $x$, a policy $\pi_{\theta_{\mathrm{old}}}$ samples a group of $n$ trajectories $\{\tau^{(k)}\}_{k=1}^{n} \sim \pi_{\theta_{\mathrm{old}}}(\cdot\mid x)$, with terminal rewards $\{R^{(k)}\}_{k=1}^{n}$. A GRPO-style update assigns each trajectory an advantage based on its reward relative to the group:
\begin{equation}
A^{(k)}
=
\frac{
R^{(k)} - \mathrm{mean}(R)
}{
\mathrm{std}(R) + \epsilon
},
\label{eq:grpo-advantage}
\end{equation}
the policy update weights each trajectory's log-probabilities by $A^{(k)}$. The same trajectory-level signal is broadcast to every action inside the rollout.

\paragraph{Policy-induced capability region.}
Let $p_\pi(c)$ denote the probability that policy $\pi$ completes the task when acting from decision context $c$. Given a reliability threshold $\eta \in (0, 1)$, we define the policy-induced capability region by:
\begin{equation}
\sC_{\eta}(\pi) = \left\{ c : p_\pi(c) \geq \eta \right\}, \label{eq:region}
\end{equation}
this region contains the decision contexts from which the current policy completes the task with probability at least $\eta$. The threshold layer around $p_\pi(c) = \eta$ is the \emph{capability boundary}. For a tolerance $\delta > 0$, the corresponding near-boundary set is:
\begin{equation}
\sN_{\eta,\delta}(\pi) = \left\{ c : \left| p_\pi(c) - \eta \right| \leq \delta \right\}, \label{eq:near-boundary}
\end{equation}
where $0 < \delta < \min\{\eta, 1 - \eta\}$. Contexts in $\sN_{\eta,\delta}(\pi)$ sit close to the current reliability threshold and are informative for expanding the capability region. The region and its boundary are induced by the current policy and need no geometric structure on the context space.

\section{Methodology}
\label{sec:method}
This section develops a method that expands the capability region $\sC_{\eta}(\pi)$ using terminal feedback alone. The method draws training contexts from near the capability boundary $\sN_{\eta,\delta}(\pi)$ and separates the learning signal into two parts. One part pushes contexts from outside the region into it, the other sharpens completion from inside the region, as illustrated in Figure~\ref{fig:overview}. We implement this expansion with a two-stage curriculum: the reverse stage uses successful reference trajectories to initialize the solvable region and expand it backward along trajectory depth, after which the forward stage broadens the capability region through natural rollouts.

\subsection{Paired advantage decomposition}
\label{sec:paired}
For a single task, we sample a rollout group of $n$ trajectories from starting contexts near the capability boundary $\sN_{\eta,\delta}(\pi)$. Let $\pi_{\theta_{\mathrm{old}}}$ denote the behavior policy used to collect the rollouts. Each rollout is divided into a prefix and a suffix. The prefix determines which decision context is reached, while the suffix determines whether the task can be completed from that context.

For the $k$-th rollout, the prefix is
$
\tau^{\mathrm{L}}_k
=
(c_{0,k},\, a_{0,k},\, o_{1,k},\, a_{1,k},\, \ldots,\, a_{\ell-1,k},\, o_{\ell,k}),
$
for $k=1,\ldots,n
$,
where $c_{0,k} \in \sN_{\eta,\delta}(\pi)$ is the starting decision context and $o_{\ell,k}$ is the prefix endpoint observation. The prefix is used to determine whether the sampled action sequence gets a context $c_\ell$ with a high probability of task completion.

The sampled prefixes are deduplicated by their endpoint context $c_\ell$. Two prefixes land in the same class if their endpoint contexts coincide, and $\mathcal{E}$ denotes the resulting set of unique endpoint contexts. Let
$
\mathcal{I}(e)
=
\{\, k : c_{\ell,k} = e \,\}
$
be the indices of prefixes with endpoint $e \in \mathcal{E}$, and let $m(e) = |\mathcal{I}(e)|$ be the number of such prefixes. Although each LLM turn contains both a \verb|<think>| segment and an executable \verb|<action>| segment, we identify duplicate endpoints using only the sequence of \verb|<action>| outputs.

For each endpoint context \(e\in\mathcal{E}\), the policy samples \(N_\mathrm{R}\geq 2\) suffixes from \(e\), where \(N_\mathrm{R}\) is the same for all endpoint contexts. The $j$-th suffix is denoted by
$
\tau^{\mathrm{R}}_{j}
=
\left(
a_{j,0},
o_{j,1},
\ldots,
a_{j,L_j-1},
o_{j,L_j}
\right)
$,
and let $R_{e,j}\in\{0,1\}$ denote its outcome. The empirical task completion value at $e$ is estimated by:
\begin{equation}
\widehat{V}(e)
=
\frac{1}{N_\mathrm{R}}
\sum_{j=1}^{N_\mathrm{R}}R_{e,j}.
\label{eq:completion-value}
\end{equation}

We compare endpoint contexts using a prefix-weighted baseline and assign the left advantage:
\begin{equation}
A^{\mathrm{L}}(e)
=
\widehat{V}(e)
-
\frac{1}{n}
\sum_{e'\in\mathcal{E}}
m(e')\,\widehat{V}(e').
\label{eq:left-adv}
\end{equation}
Thus, $A^{\mathrm{L}}(e)$ measures whether prefixes reaching endpoint $e$ have a higher empirical task-completion value than the prefix-weighted average over all endpoints. The suffixes associated with the same endpoint context are compared separately. For suffix $j$ sampled from $e$, define the right advantage by:
\begin{equation}
A^{\mathrm{R}}(e,j)
=
R_{e,j}
-
\frac{1}{N_\mathrm{R}-1}
\sum_{j'\neq j}R_{e,j'}.
\label{eq:right-adv}
\end{equation}
This signal evaluates a suffix relative to the other suffixes sampled from the same endpoint context. When all suffixes associated with $e$ have the same outcome, the right signal vanishes, while $e$ can still contribute through the left signal in Eq.~\ref{eq:left-adv}.

The paired decomposition therefore separates two aspects of capability expansion. The left signal evaluates which prefix executions reach contexts with higher completion probabilities, whereas the right evaluates how the task should be completed after a particular context has been reached.

To realize this decomposition in the policy update, each prefix is completed jointly with a suffix sampled from its endpoint. For prefix $\tau_k^{\mathrm{L}}$, let $e_k=c_{\ell,k}$ denote its endpoint context and independently sample $j_k$ uniformly from $\{1,\ldots,N_R\}$. The selected suffix $\tau_k^{\mathrm{R}}=\tau_{e_k,j_k}^{\mathrm{R}}$ forms a completed trajectory $\tau_k=\tau_k^{\mathrm{L}}\oplus\tau_k^{\mathrm{R}}$, where $\oplus$ denotes concatenation. With the estimated advantages held fixed, the policy optimization objective is:
\begin{equation}
\begin{aligned}
\mathcal{J}(\theta)
=
\mathbb{E}\Bigg[
\frac{1}{n}\sum_{k=1}^{n}
\Bigg(
\frac{1}{\ell}
\sum_{t=0}^{\ell-1}
\min\Big[
\rho_{t,k}(\theta)A^{\mathrm{L}}(e_k),
\operatorname{clip}\big(\rho_{t,k}(\theta),1-\epsilon,1+\epsilon\big)
A^{\mathrm{L}}(e_k)
\Big]
\\
+
\frac{1}{T_k-\ell}
\sum_{t=\ell}^{T_k-1}
\min\Big[
\rho_{t,k}(\theta)A^{\mathrm{R}}(e_k,j_k),
\operatorname{clip}\big(\rho_{t,k}(\theta),1-\epsilon,1+\epsilon\big)
A^{\mathrm{R}}(e_k,j_k)
\Big]
\Bigg)
\Bigg].
\label{eq:surrogate}
\end{aligned}
\end{equation}
where $\rho_{t,k}(\theta)
=
\dfrac{\pi_\theta(a_{t,k}\mid c_{t,k})}
       {\pi_{\theta_{\mathrm{old}}}(a_{t,k}\mid c_{t,k})}$
is the importance sampling ratio, $\epsilon$ is the clipping parameter, and $T_k$ is the number of actions in the completed trajectory $\tau_k$. The expectation is over task sampling, rollout collection under $\pi_{\theta_{\mathrm{old}}}$, and suffix selection. The two terms apply clipped relative advantages to the prefix and suffix actions, respectively. The two-stage curriculum uses the same objective and differs in their starting contexts and prefix lengths used to construct these trajectories.

\subsection{Two-stage curriculum: reverse then forward}
\label{sec:curriculum}
The paired advantage update samples starting contexts from the near-boundary set $\sN_{\eta,\delta}(\pi)$. The curriculum constructs these contexts in two stages. The reverse stage expands $\sC_{\eta}(\pi)$ backward along successful reference trajectories, while the forward stage extends it to contexts encountered in the policy's own rollouts.

\paragraph{Reverse stage: expanding backward along references.}
\label{sec:reverse}
The reverse stage operates on training tasks drawn from $\mathcal{D}$. For each task, we assume a successful reference trajectory
$
\tau^*
=
\left\{
(o_t^*,a_t^*)
\right\}_{t=0}^{T-1}
$. The reverse stage starts from a context $ c_{T-1}^* = (\,o_{\leq T-1}^*,\, a_{< T-1}^*, x)$ near the end of $\tau^*$, where the remaining effective horizon is short, and progressively moves toward the original task start.

Let $r_0=T-1$ denote the initial position of the reverse curriculum. At stage $i$, we select the current near-boundary context $c_i=c_{r_i}^*$, where
$
r_i=r_0-i d
$
and $d$ is the fixed backward stride along the reference trajectory. We sample prefixes of length $\ell_i= d+2$ from $c_i$, group them by their endpoint contexts, and apply the paired advantage decomposition to the resulting groups.

The update continues until the estimated success probability at $c_i$ reaches the reliability threshold
$
p_\pi(c_i)\geq\eta
$,
after which $c_i$ is treated as belonging to $\sC_{\eta}(\pi)$. The curriculum then proceeds to the earlier reference context indexed by $r_{i+1}$. The reverse stage terminates when the original task start $c_0^*=(o_0^*,x)$ is included in $\sC_{\eta}(\pi)$ for most training tasks. This produces a policy that can reliably solve the training tasks from their original starts, while its capability region may still be concentrated around the reference trajectories.

\paragraph{Forward stage: extending into the policy's own distribution.}
\label{sec:forward}
The forward stage uses the now-solvable original context $c_0=(o_0,x)$ as a rollout anchor and extends $\sC_{\eta}(\pi)$ to contexts generated by the policy itself. At forward stage $i$, prefixes are sampled from $c_0$ with length $\ell_i$, initialized at $\ell_0=d+2$ to match the final reverse-stage horizon and then increased according to a monotone schedule, for example $\ell_i=d+2+i\Delta\ell$ with $\Delta\ell>0$. The resulting prefixes may reach contexts outside the reference trajectories, and these endpoint contexts are then trained with the paired advantage update and incorporated into the expanding capability region.

The two stages therefore use the same update but different prefix distributions. The reverse stage samples prefixes from successive contexts along successful references and moves the capability boundary toward earlier task contexts. The forward stage samples prefixes from policy-generated rollouts starting at $c_0$ and expands the boundary to contexts outside the support of the references.

\paragraph{Curriculum progression.}
In the reverse stage, successful reference trajectories provide an ordered sequence of candidate starting contexts. Once the current start reaches the reliability threshold $\eta$, moving backward along the reference trajectory tends to place the next start near the expanding capability boundary, as illustrated in Figure~\ref{fig:decay}. We therefore use these starts as practical near-boundary contexts without explicitly computing $\sN_{\eta,\delta}(\pi)$ or specifying a numerical $\delta$. In the forward stage, we monitor the mean number of distinct prefix endpoints $|\mathcal{E}|$ per rollout group. When this mean stabilizes over successive updates, indicating that endpoint diversity at the current prefix length has plateaued, we advance the curriculum and increase the prefix length by $\Delta\ell$.Increasing the prefix length lets the policy search for endpoint contexts $e$ that may lie across the capability boundary.

\section{Experiments}
\label{sec:exp}

\subsection{Experimental Setup}
\label{sec:exp-setup}
\paragraph{Benchmarks.}
We evaluate on ALFWorld \citep{shridhar2021} and WebShop \citep{yao2022}. \textbf{ALFWorld} comprises 3{,}827 embodied household tasks across six categories (\textsc{Pick}, \textsc{Look}, \textsc{Clean}, \textsc{Heat}, \textsc{Cool}, and \textsc{Pick2}). Given a textual goal, the agent completes each task through multi-turn interaction with the environment. \textbf{WebShop} simulates online shopping through an HTML interface in which the agent searches for products, navigates pages, and makes a purchase based on a user instruction. It contains over 1.1 million products and 12{,}000 user instructions.

\paragraph{Baselines.}
We compare against closed-source LLMs, prompting agents, and RL-trained agents as three families of baselines. \textbf{Closed-source models} are GPT-4o \citep{achiam2023} and Gemini-2.5-Pro \citep{team2023}, representing strong general-purpose reasoning. \textbf{Prompting agents} include Qwen2.5-Instruct, ReAct \citep{yao2023} and Reflexion \citep{shinn2023}, relying on prompts without parameter updates. \textbf{RL-based training methods} cover PPO \citep{schulman2017}, RLOO \citep{kool2019,ahmadian2024}, and GRPO \citep{shao2024} under trajectory-level supervision, alongside methods that address long-horizon credit assignment through step-level grouping GiGPO\citep{feng2025} and reverse replay CRPS\citep{zhang2026}.

\paragraph{Implementation Details.}
We adopt Qwen2.5-1.5B-Instruct and Qwen2.5-7B-Instruct \citep{qwen2025} as base models. All RL methods use the same backbone and training settings, with 16 tasks per update, 8 rollouts per task. Agents generate reasoning within \verb|<think>...</think>| tags before emitting an action within \verb|<action>...</action>| tags. Episodes are capped at 50 turns in ALFWorld and 15 in WebShop. Our method samples $n=8$ prefixes per task and 2 suffixes per unqiue endpoint. All training runs on 4 NVIDIA H100 GPUs. More training setups and hyperparameter details are provided in Appendix~\ref{app:impl}.

\subsection{Main Results}
\label{sec:exp-res}
Table~\ref{tab:main} reports the result of our method at both model scales, compared with prompting and representative reinforcement learning methods under the matched evaluation protocol. RL training improves substantially over prompting alone on both benchmarks, confirming the difficulty of learning long-horizon interaction from prompts alone. On ALFWorld, our method reaches $90.1\%$ overall success with Qwen2.5-1.5B-Instruct and $96.1\%$ with Qwen2.5-7B-Instruct. These results exceed GRPO baseline by $17.3$ and $18.5$ percentage points, respectively, and the competitive GiGPO by $3.4$ and $5.3$ points. It leads in five of the six task categories at each scale. The exceptions are Clean at 1.5B ($84.2\%$ versus GiGPO's $94.8\%$) and Cool at 7B ($85.9\%$ versus $89.3\%$). The overall gains over GRPO and GiGPO indicate that the method remains effective across tasks with different interaction requirements, rather than relying on improvements in a single category.

\begin{table}[t]
\centering
\caption{Performance on ALFWorld and WebShop. We report success rates (\%) for ALFWorld and task scores and success rates (\%) for WebShop. Subscripts denote standard deviations where available. For our method, they are computed over three evaluation seeds using a fixed checkpoint. Baseline results are taken from GiGPO~\citep{feng2025} and CRPS~\citep{zhang2026}. Bold indicates the best mean in each column within a model scale.}
\label{tab:main}
\small
\setlength{\tabcolsep}{3.6pt}
\renewcommand{\arraystretch}{1.08}
\resizebox{\linewidth}{!}{%
\begin{tabular}{@{}c l l ccccccc cc@{}}
\toprule
& \multirow{2}{*}{\textbf{Type}} & \multirow{2}{*}{\textbf{Method}}
& \multicolumn{7}{c}{\textbf{ALFWorld}}
& \multicolumn{2}{c}{\textbf{WebShop}} \\
\cmidrule(lr){4-10}\cmidrule(lr){11-12}
& & & Pick & Look & Clean & Heat & Cool & Pick2 & All & Score & Succ. \\
\midrule
\multicolumn{12}{l}{\emph{Closed-Source Models}} \\
& Prompting & GPT-4o
& 75.3 & 60.8 & 31.2 & 56.7 & 21.6 & 49.8 & 48.0 & 31.8 & 23.7 \\
& Prompting & Gemini-2.5-Pro
& 92.8 & 63.3 & 62.1 & 69.0 & 26.6 & 58.7 & 60.3 & 42.5 & 35.9 \\
\midrule
\multirow{9}{*}{\rotatebox{90}{\textbf{1.5B}}}
& Prompting & Qwen2.5
& 5.9 & 5.5 & 3.3 & 9.7 & 4.2 & 0.0 & 4.1 & 23.1 & 5.2 \\
& Prompting & ReAct
& 17.4 & 20.5 & 15.7 & 6.2 & 7.7 & 2.0 & 12.8 & 40.1 & 11.3 \\
& Prompting & Reflexion
& 35.3 & 22.2 & 21.7 & 13.6 & 19.4 & 3.7 & 21.8 & 55.8 & 21.9 \\
\cdashline{2-12}
& RL Training & PPO
& $64.8\pmstd{3.5}$
& $40.5\pmstd{6.9}$
& $57.1\pmstd{4.9}$
& $60.6\pmstd{6.6}$
& $46.4\pmstd{4.0}$
& $47.4\pmstd{1.9}$
& $54.4\pmstd{3.1}$
& $73.8\pmstd{3.0}$
& $51.5\pmstd{2.9}$ \\
& RL Training & RLOO
& $88.3\pmstd{3.0}$
& $52.8\pmstd{8.6}$
& $71.0\pmstd{5.9}$
& $62.8\pmstd{8.7}$
& $66.4\pmstd{5.5}$
& $56.9\pmstd{4.7}$
& $69.7\pmstd{2.5}$
& $73.9\pmstd{5.6}$
& $52.1\pmstd{6.7}$ \\
& RL Training & GRPO
& $85.3\pmstd{1.5}$
& $53.7\pmstd{8.0}$
& $84.5\pmstd{6.8}$
& $78.2\pmstd{7.9}$
& $59.7\pmstd{5.0}$
& $53.5\pmstd{5.6}$
& $72.8\pmstd{3.6}$
& $75.8\pmstd{3.5}$
& $56.8\pmstd{3.8}$ \\
& RL Training & CRPS
& 91.0 & 72.0 & 88.0 & 90.0 & 73.0 & 76.0 & 83.0 & 73.0 & 61.0 \\
& RL Training & GiGPO
& $94.4\pmstd{5.9}$
& $67.5\pmstd{4.6}$
& $\mathbf{94.8}\pmstd{3.8}$
& $94.4\pmstd{7.8}$
& $79.8\pmstd{4.7}$
& $76.4\pmstd{5.4}$
& $86.7\pmstd{1.7}$
& $83.1\pmstd{1.6}$
& $65.0\pmstd{3.2}$ \\
\rowcolor{cyan!20}
& RL Training & \texttt{\textbf{Ours}}
& $\mathbf{97.9}\pmstd{3.7}$
& $\mathbf{97.2}\pmstd{4.8}$
& $84.2\pmstd{4.23}$
& $\mathbf{100.0}\pmstd{0.0}$
& $\mathbf{80.3}\pmstd{4.0}$
& $\mathbf{86.4}\pmstd{4.6}$
& $\mathbf{90.1}\pmstd{2.0}$
& $\mathbf{88.6}\pmstd{1.5}$
& $\mathbf{80.2}\pmstd{3.9}$ \\
\midrule
\multirow{9}{*}{\rotatebox{90}{\textbf{7B}}}
& Prompting & Qwen2.5
& 33.4 & 21.6 & 19.3 & 6.9 & 2.8 & 3.2 & 14.8 & 26.4 & 7.8 \\
& Prompting & ReAct
& 48.5 & 35.4 & 34.3 & 13.2 & 18.2 & 17.6 & 31.2 & 46.2 & 19.5 \\
& Prompting & Reflexion
& 62.0 & 41.6 & 44.9 & 30.9 & 36.3 & 23.8 & 42.7 & 58.1 & 28.8 \\
\cdashline{2-12}
& RL Training & PPO
& $92.3\pmstd{4.0}$
& $64.0\pmstd{8.4}$
& $92.5\pmstd{2.4}$
& $89.5\pmstd{7.0}$
& $80.3\pmstd{2.0}$
& $68.8\pmstd{8.3}$
& $80.4\pmstd{2.7}$
& $81.4\pmstd{3.1}$
& $68.7\pmstd{5.1}$ \\
& RL Training & RLOO
& $87.6\pmstd{4.3}$
& $78.2\pmstd{8.3}$
& $87.3\pmstd{5.8}$
& $81.3\pmstd{7.6}$
& $71.9\pmstd{5.2}$
& $48.9\pmstd{8.4}$
& $75.5\pmstd{4.6}$
& $80.3\pmstd{3.2}$
& $65.7\pmstd{4.0}$ \\
& RL Training & GRPO
& $90.8\pmstd{5.1}$
& $66.1\pmstd{6.7}$
& $89.3\pmstd{5.4}$
& $74.7\pmstd{6.9}$
& $72.5\pmstd{5.4}$
& $64.7\pmstd{7.3}$
& $77.6\pmstd{5.2}$
& $79.3\pmstd{2.8}$
& $66.1\pmstd{3.7}$ \\
& RL Training & CRPS
& 91.0 & 80.0 & 86.0 & 83.0 & 80.0 & 71.0 & 86.0 & 82.0 & 74.0 \\
& RL Training & GiGPO
& $97.7\pmstd{1.6}$
& $82.7\pmstd{7.9}$
& $98.8\pmstd{1.6}$
& $83.7\pmstd{7.2}$
& $\mathbf{89.3}\pmstd{8.2}$
& $79.2\pmstd{6.6}$
& $90.8\pmstd{1.3}$
& $84.4\pmstd{2.9}$
& $72.8\pmstd{3.2}$ \\
\rowcolor{cyan!20}
& RL Training & \texttt{\textbf{Ours}}
& $\mathbf{100.0}\pmstd{0.0}$
& $\mathbf{100.0}\pmstd{0.0}$
& $\mathbf{100.0}\pmstd{0.0}$
& $\mathbf{100.0}\pmstd{0.0}$
& $85.9\pmstd{6.0}$
& $\mathbf{92.4}\pmstd{2.6}$
& $\mathbf{96.1}\pmstd{1.4}$
& $\mathbf{92.5}\pmstd{1.1}$
& $\mathbf{82.6}\pmstd{1.2}$ \\
\bottomrule
\end{tabular}%
}
\end{table}

On WebShop, our method achieves a score of $88.6$ and a success rate of $80.2\%$ at 1.5B, compared with $83.1$ and $65.0\%$ for GiGPO. At 7B, it reaches $92.5$ and $82.6\%$. The score exceeds GiGPO by $8.1$ points, while the success rate exceeds CRPS by $8.6$ percentage points. Our method leads on both WebShop metrics at both scales. The larger gains in success rate than in task score are consistent with more agents completing the full sequence of search, navigation, and product-selection decisions. The ablation results below examine the contribution of paired credit assignment to these gains.

\subsection{Analysis}
\label{sec:analysis}

\begin{figure}[t]
\centering
\begin{subfigure}[t]{0.48\linewidth}
  \centering
  \includegraphics[width=\linewidth]{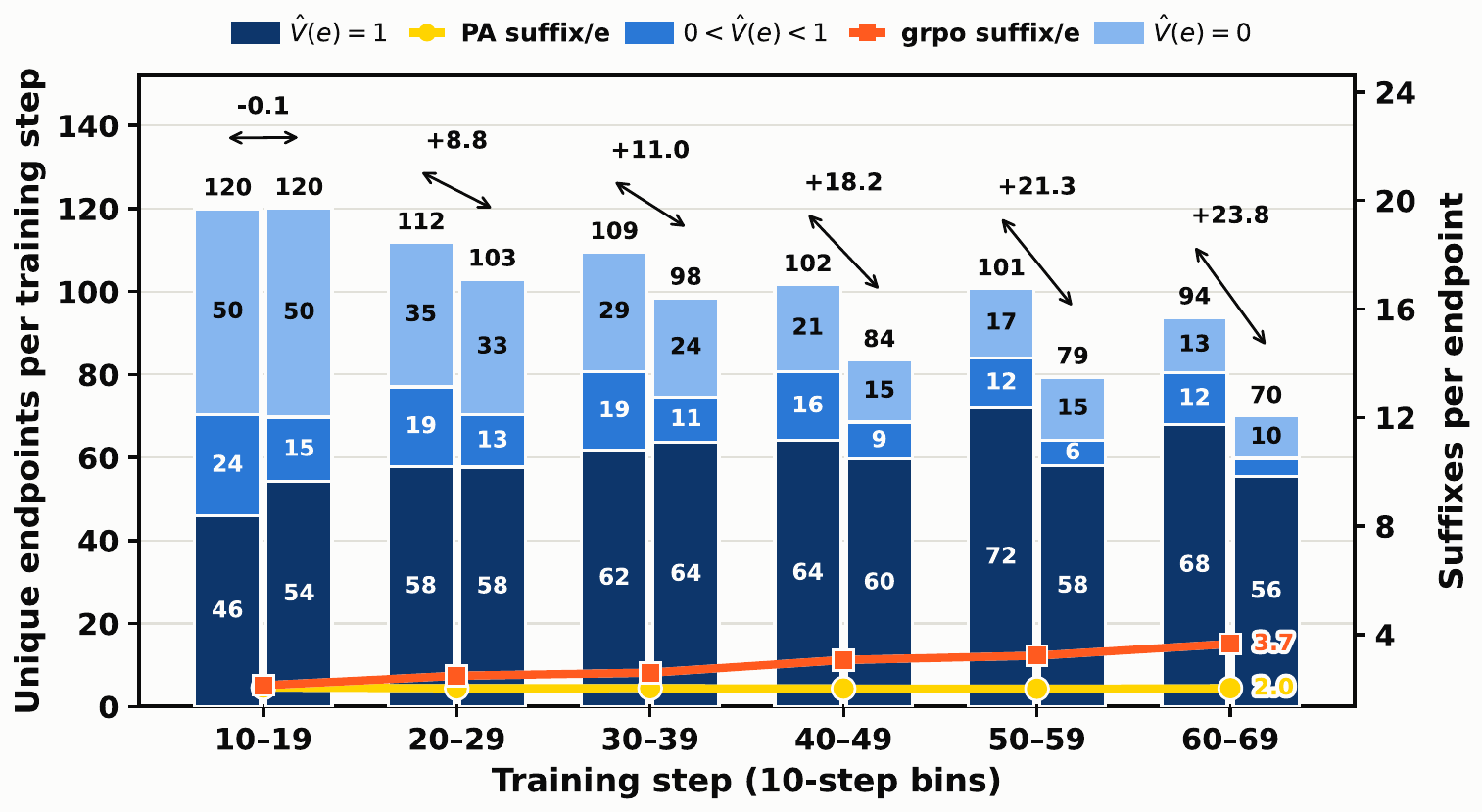}
  \caption{}
  \label{fig:alfworld-7b-endpoint}
\end{subfigure}\hfill
\begin{subfigure}[t]{0.48\linewidth}
  \centering
  \includegraphics[width=\linewidth]{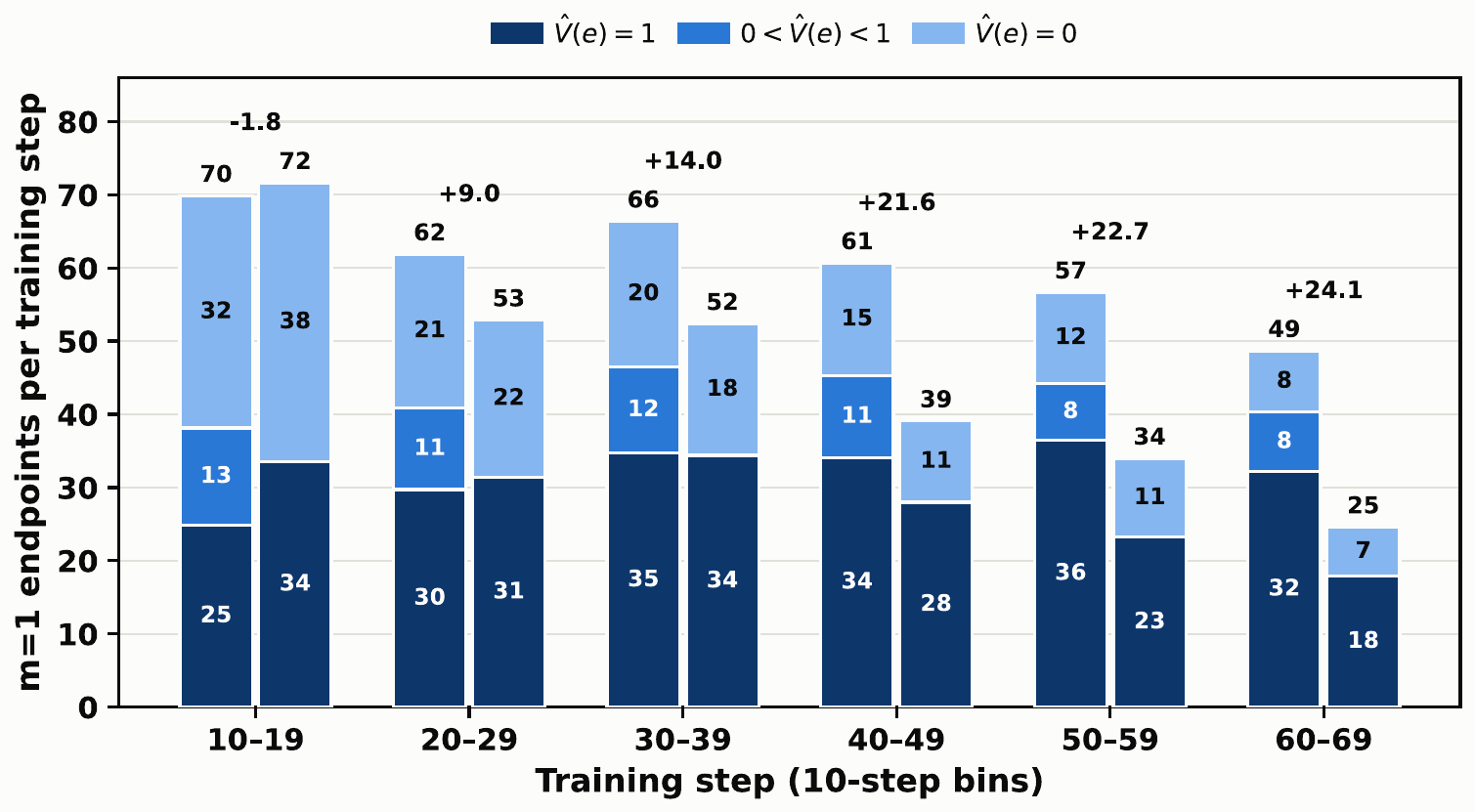}
  \caption{}
  \label{fig:alfworld-7b-singleton}
\end{subfigure}
\caption{\textbf{Prefix endpoint coverage during ALFWorld training with Qwen2.5-7B-Instruct.}
In two panels, the left bar of each pair represents PA and the right bar represents GRPO. Both methods are initialized from the same model checkpoint and trained on the same dataset.  PA samples exactly two suffixes from each unique endpoint, whereas GRPO observes one continuation for each prefix reaching that endpoint, so its suffix count is determined by $m(e)$.
(a) Bars show the mean number of distinct prefix endpoints per training step, grouped by  value $\widehat{V}(e)$.Curves show the mean number of suffixes per endpoint.
(b) Endpoint counts restricted to those reached by exactly one prefix, $m(e)=1$. PA's two suffixes can reveal mixed outcomes $\widehat{V}(e)=0.5$ at these singleton endpoints. }
\label{fig:alfworld-7b-endpoint-pair}
\end{figure}

\begin{wrapfigure}{r}{0.38\linewidth}
\centering
\includegraphics[width=\linewidth]{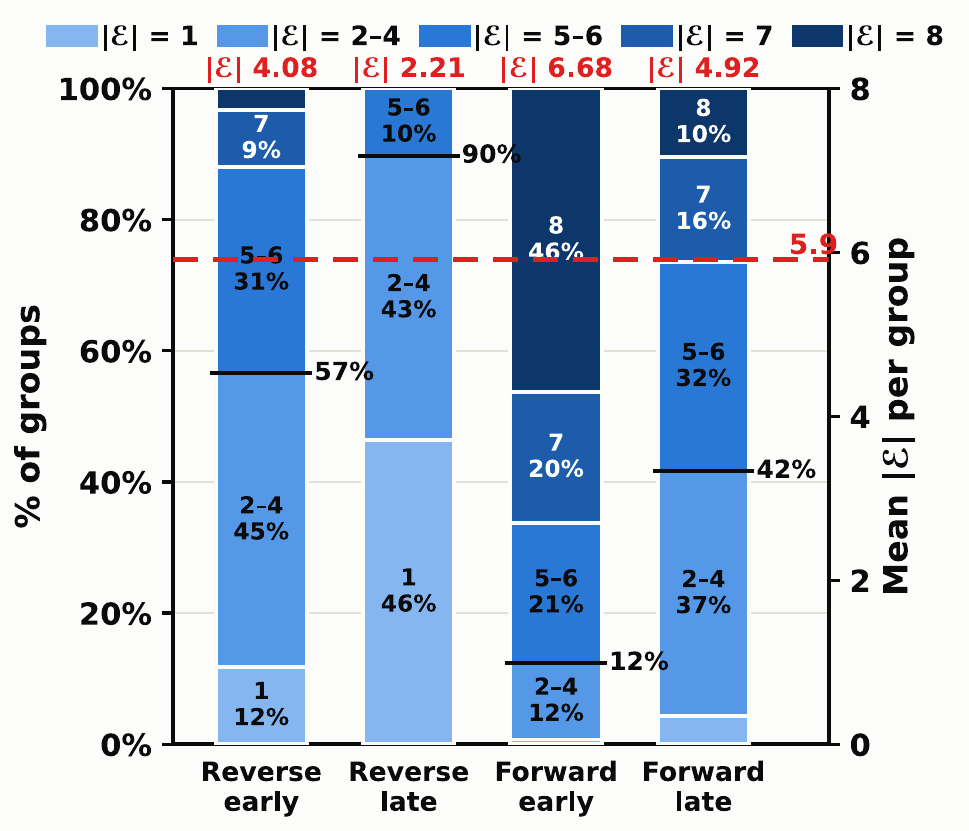}
\caption{Distribution of the number of unique prefix endpoint contexts $|\mathcal{E}|$ reached per training on Webshop.}
\label{fig:group-endpoint}
\end{wrapfigure}

\textbf{PA decomposition expands capability region.} Figure~\ref{fig:alfworld-7b-endpoint} analyzes the prefix endpoints reached during ALFWorld training with Qwen2.5-7B-Instruct. The bars show the mean number of distinct endpoint contexts \(e\) per training step, grouped by  value \(\widehat{V}(e)\), and the curves show the mean number of observed suffixes per endpoint. PA has $0.1$ fewer endpoints per step than GRPO in the first bin but $23.8$ more in the final bin, reaching $93.8$ versus $70.0$ endpoints per step. including \(68.0\) versus \(55.5\) with \(\widehat{V}(e)=1\). Meanwhile, PA maintains a fixed budget of $N_{\mathrm{R}}=2$ suffixes per endpoint by design (Eq.~\ref{eq:completion-value}), in contrast to GRPO's observed increase from $2.1$ to $3.7$. This pattern is consistent with a winner-takes-all effect under GRPO: trajectory-level advantages reinforce prefixes associated with successful rollouts, so sampled prefixes may converge on fewer endpoint contexts. PA instead evaluates each distinct endpoint with two suffixes and separates prefix selection from continuation quality, keeping more potentially solvable endpoints available for learning.

Figure~\ref{fig:alfworld-7b-singleton} offers a clearer view of this divergence by isolating endpoints reached by exactly one prefix, \(m(e)=1\). For these singleton endpoints, GRPO observes one binary outcome and cannot assign an intermediate empirical value, PA's two suffixes produce \(\widehat{V}(e)=0.5\) when one succeeds and the other fails, enabling the within-endpoint comparison in its right advantage (Eq.~\ref{eq:right-adv}). From the first to the final bin, PA's singleton coverage falls from about \(70\) to \(48.7\) endpoints per step, yet its value-\(1\) count rises from about \(25\) to \(32.2\), while its value-\(0.5\) count falls from about \(13\) to \(8.1\). GRPO loses both singleton coverage (about \(72\) to \(24.6\)) and value-\(1\) endpoints (about \(34\) to \(17.9\)). These contrasting trends are consistent with PA's within-endpoint comparisons helping partially solvable contexts become more reliably solvable, while GRPO's observed endpoint distribution contracts in both breadth and high-value count.

\textbf{Why reverse, then forward?} Figure~\ref{fig:group-endpoint} reports the distribution of $|\mathcal{E}|$, the number of unique prefix endpoint contexts reached per training group, across four consecutive training stages on WebShop. The reverse stage drives the mean $|\mathcal{E}|$ from $4.08$ down to $2.21$ with the $|\mathcal{E}|{=}1$ fraction reaching $46\%$ by late reverse, evidence that the policy has collapsed most rollouts onto a single reachable endpoint per group. Through this collapse, the reverse stage establishes the initial capability region, with tight concentration on the reference trajectory yielding a small but reliable solvable region. The forward stage takes this region as its anchor and pushes the mean $|\mathcal{E}|$ back to $6.68$ in early forward, with $66\%$ of groups at $|\mathcal{E}|\ge 7$, before settling at $4.92$ in late forward. Read across the four stages, the capability region is first anchored then expanded. The reverse stage  establishes it along the reference trajectory, and the forward stage continues expansion into the off-reference states the policy now visits. 

\subsection{Ablation Study}
\label{sec:abl-sty}
We conduct an ablation study to evaluate the effectiveness of each component in our method. As illustrated in Table~\ref{tab:ablation-webshop}, the two curriculum stages and paired advantages both contribute to the performance on WebShop using Qwen2.5-1.5B-Instruct. The reverse-only configuration baseline without paired advantages achieves a success rate of 68.8\%. Adding the forward stage increases performance to 74.7\%. Incorporating paired advantages into the reverse stage further improves the success rate to 76.6\%, and in both curriculum stages achieves the best performance of 80.20\%, which is 11.4 points higher than the reverse-only baseline. The gains from applying PA in both stages support the benefit of separately crediting prefix actions for the endpoints they reach and suffix actions for their outcomes from those endpoints.
\vspace{-6pt}
\begin{table}[t]
\centering
\caption{Ablation study on WebShop using Qwen2.5-1.5B-Instruct.}
\label{tab:ablation-webshop}
\vspace{-4pt}
\footnotesize
\setlength{\tabcolsep}{4.5pt}
\renewcommand{\arraystretch}{0.95}
\begin{tabular}{@{}lcc c@{}}
\toprule
\textbf{Curriculum} & \textbf{Rev.\ PA} & \textbf{Fwd.\ PA} & \textbf{Succ.\ (\%)} \\
\midrule
Reverse only & -- & -- & 68.8 \\
Reverse + Forward & -- & -- & 74.7 \\
Reverse + Forward & $\checkmark$ & -- & 76.6 \\
Reverse + Forward & $\checkmark$ & $\checkmark$ & \textbf{80.2} \\
\bottomrule
\end{tabular}
\vspace{-8pt}
\end{table}

\subsection{Training Efficiency and Cost Analysis}
\label{sec:cost-anlysis}

\begin{wrapfigure}{r}{0.595\linewidth}
\centering
\begin{subfigure}[t]{0.2938\textwidth}
  \centering
  \includegraphics[width=\linewidth]{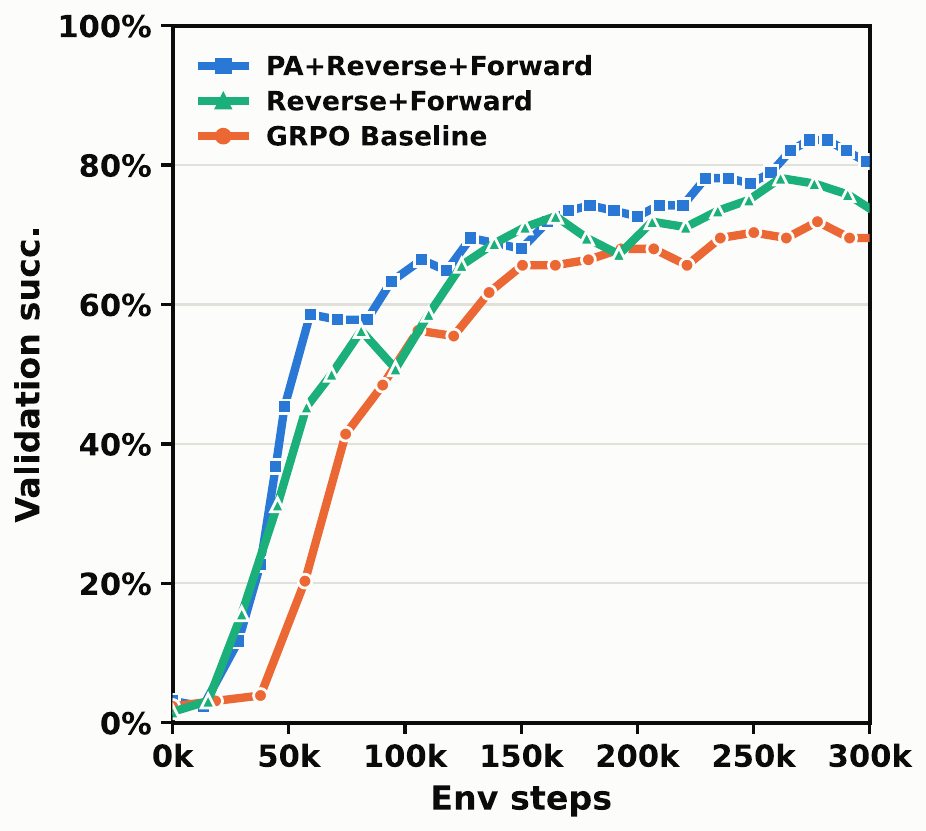}
  \caption{}
  \label{fig:cost-vs-val}
\end{subfigure}\hspace{0.6em}%
\begin{subfigure}[t]{0.2849\textwidth}
  \centering
  \includegraphics[width=\linewidth]{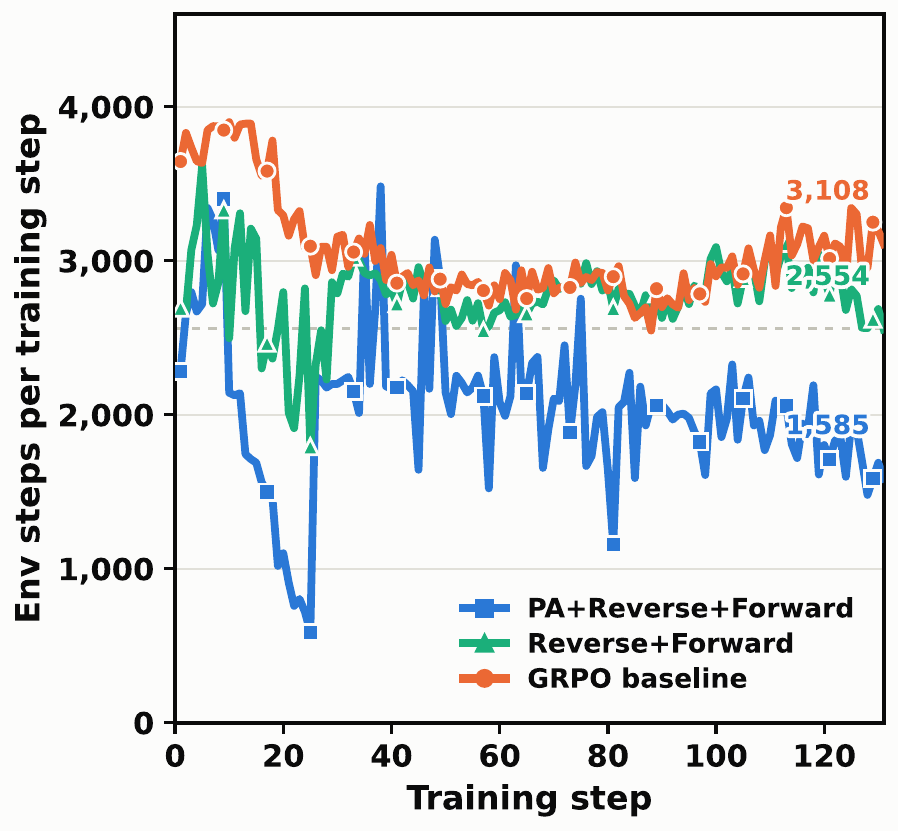}
  \caption{}
  \label{fig:step-cost}
\end{subfigure}
\caption{(a) Validation success against cumulative model-generated environment steps. (b) Environment steps consumed per training step for each arm.}
\label{fig:webshop-cost}
\end{wrapfigure}
Figure~\ref{fig:cost-vs-val} shows that PA achieves higher validation success at matched environment-interaction budgets, despite reaching about $5.9$ distinct prefix endpoints per group in Figure~\ref{fig:group-endpoint} and sampling two suffixes from each nonterminal endpoint. Figure~\ref{fig:step-cost} helps explain why the additional branching does not necessarily increase interaction cost. Reverse training starts near the ends of successful references, keeping the remaining rollouts short. By late reverse, eight sampled prefixes reach only $2.21$ distinct endpoints per group on average, limiting the number of endpoints that require suffix sampling. In forward training, endpoint diversity increases as PA explores contexts beyond the references. The reverse-trained policy helps avoid long, unproductive continuations from these endpoints, allowing PA to explore more contexts without a proportional increase in environment steps. The occasional cost spikes are consistent with batches reaching more endpoints, but PA generally uses fewer environment steps per training step than the baselines.

\section{Related Work}
\label{sec:related}
\paragraph{Reinforcement Learning for Long-Horizon LLM Agents.}
LLM agents extend language models to long-horizon sequential decisions through repeated interaction with environments. The paradigm has evolved from static text generation, through single-turn tool use \citep{schick2023}, to multi-turn sequential decision making, exemplified by ReAct and Reflexion \citep{yao2023,shinn2023}. In LLM training, the RLHF pipeline paired PPO with a learned value function \citep{ouyang2022,schulman2017}, while RLOO and GRPO removed the need for a separate critic \citep{ahmadian2024,shao2024}, and AGILE and WebRL subsequently applied reinforcement learning to agent settings \citep{feng2024,qi2025}. Recent work has brought online RL to long-horizon agents through dedicated multi-turn environments and scalable training systems such as RAGEN, AgentGym-RL, and SkyRL-Agent \citep{wang2025,xi2025,cao2025}. These efforts extend RL training to complete agent episodes across search, software engineering, and mobile GUI tasks \citep{kalyan2025,cao2025,xu2026}.

\paragraph{Credit Assignment.}
In sparse reward settings, long-horizon agents typically receive supervision only from a terminal outcome, which makes credit assignment over intermediate decisions difficult. Process reward models and search-based approaches introduce intermediate signals, but require additional manual annotations and expensive search computation, respectively \citep{lightman2024,hao2023}. Later agentic RL methods such as GiGPO and HGPO derive finer grained comparisons from repeated states across rollouts without externally labeled process rewards \citep{feng2025,he2026}, whereas STAPO derives a selective step-level signal via normalized entropy \citep{stapo2026}. More recent approaches modify the rollout structure itself by aggregating trajectories into transition graphs or introducing structured branching and information guided exploration, as in GraphGPO, SPARK, Tree GRPO, and AEPO \citep{cheng2026,wu2026,ji2025,dong2025}. E3-TIR takes a complementary approach by reusing a common continuation to compare prefixes, allowing downstream outcomes to provide evidence about prefix quality \citep{guo2026}. Taken together, these studies suggest that long-horizon credit assignment depends on how rollout structure determines which decisions share, inherit, or receive separately estimated learning signals.

\paragraph{Curriculum Learning.}
Curriculum learning structures the training distribution so that competence acquired on simpler subproblems supports learning on more difficult ones \citep{bengio2009}. In reinforcement learning and robot control, this principle has been implemented by adapting restart states or task sequences around near-goal and demonstrated states \citep{tavakoli2019,florensa2017,resnick2018,ivanovic2019,wohlke2020,sharma2021}. In agentic RL, curricula modify task distribution, the available supervision, or the interaction environment to regulate the difficulty of multi-turn interaction \citep{tan2025,tzannetos2025,xia2025}. Recent work derives curricula directly from interaction outcomes, using failed attempts to generate new tasks and successful trajectories to identify informative restart states \citep{qi2025,zhang2026}. Our method builds on curriculum learning by first expanding the training start backward along successful reference trajectories and then broadening training to states generated by the policy itself.

\section{Conclusion}
\label{sec:conclusion}

We train long-horizon agents from sparse terminal rewards by progressively expanding the policy's capability region. Paired Advantage separately credits prefixes for reaching more solvable contexts and suffixes for task completion with a reverse-to-forward curriculum moves from successful references to policy-generated contexts. The method outperforms strong baselines on ALFWorld and WebShop at both model scales.

{\bf Limitations.} Reverse training requires successful references, and suffix sampling requires restarting from intermediate contexts. These requirements restrict use when references or reliable resets are unavailable. If all sampled suffixes fail, both advantages vanish. Starting near known solutions reduces this risk, but expansion may stall when solvable contexts remain beyond the reach of sampled prefixes.

\newpage
\subsection*{AI use statement}
In this work, we used generative AI tools to assist with code implementation and to improve the grammar, clarity, and readability of the manuscript. We have not used generative AI tools for generating research proposals, methodological designs, experimental plans, or scientific conclusions. We have reviewed and tested all AI-assisted code, and all reported results come from running the verified implementation by 2 authors. We also reviewed all AI-assisted writing to ensure it accurately reflects the methods, results, and claims of this work. We take full responsibility for the final content of this paper, including all text, claims, code, and other artifacts produced with the assistance of generative AI.

\bibliography{iclr2027_conference}
\bibliographystyle{iclr2027_conference}

\clearpage

\appendix
\raggedbottom

\section{Algorithm}
\label{app:algorithms}
Algorithm~\ref{alg:depth-before-breadth} summarizes the training procedure. Endpoint grouping depends on the full decision context, which includes both the task and the interaction history. As a result, the same observation does not by itself guarantee the same endpoint. For each distinct endpoint, we sample $N_{\mathrm R}$ suffixes once and allow every prefix reaching that endpoint to reuse them. Each iteration thus samples $|\mathcal E|N_{\mathrm R}$ suffixes instead of $nN_{\mathrm R}$. In the left advantage, the factor $m(e)$ preserves the number of prefixes reaching each endpoint in the group baseline. Deduplication then reduces suffix sampling without changing how prefixes are weighted. The reverse and forward stages share the same paired advantage update. The transition between the two stages is determined by the estimated completion reliability at the original task starts.

\begin{algorithm}[H]
\caption{Training LLM agents with Depth Before Breadth}
\label{alg:depth-before-breadth}
\begin{algorithmic}[1]
\Require Policy $\pi_\theta$, tasks $\mathcal D$, successful references
         $\{\tau_x^*\}$, group size $n$, suffix count $N_{\mathrm R}$,
         reverse stride $d$, threshold $\eta$, forward horizons $\{\ell_i\}$
\Ensure Updated policy $\pi_\theta$
\State Initialize the curriculum: $q\gets\mathrm{reverse}$; $i\gets0$
\For{each training iteration}
    \State Update the old policy model $\theta_{\mathrm{old}}\gets\theta$ and sample task $x\sim\mathcal D$
    \State \textit{// Select the curriculum anchor}
    \If{$q=\mathrm{reverse}$}
        \State Move along the reference:
               $r_i\gets\max\{T_x-1-id,0\}$;
               $(c,\ell)\gets(c_{r_i}^*,d+2)$
    \Else
        \State Start from the original context:
               $(c,\ell)\gets(c_0,\ell_i)$
    \EndIf
    \State \textit{// Sample prefixes and group shared endpoints}
    \For{$k=1,\ldots,n$}
        \State Sample a prefix:
               $\tau_k^{\mathrm L}
               \sim\pi_{\theta_{\mathrm{old}}}(\cdot\mid c)$;
               $e_k\gets c_{\ell,k}$
    \EndFor
    \State Group endpoints:
           $\mathcal E\gets\{e_k\}_{k=1}^n$;
           $\mathcal I(e)\gets\{k:e_k=e\}$;
           $m(e)\gets|\mathcal I(e)|$
    \State \textit{// Sample suffixes and compute paired advantages}
    \For{each $e\in\mathcal E$}
        \State Sample continuations:
               $\{(\tau_{e,j}^{\mathrm R},R_{e,j})\}
               _{j=1}^{N_{\mathrm R}}$
               from $e$ under $\pi_{\theta_{\mathrm{old}}}$
        \State Estimate endpoint value $\widehat V(e)$
               by Eq.~\ref{eq:completion-value}
        \State Compute suffix advantages $A^{\mathrm R}(e,j)$
               by Eq.~\ref{eq:right-adv}
    \EndFor
    \State Compute prefix advantages $A^{\mathrm L}(e)$
           by Eq.~\ref{eq:left-adv}
    \For{$k=1,\ldots,n$}
        \State Select a suffix:
               $j_k\sim\operatorname{Unif}
               \{1,\ldots,N_{\mathrm R}\}$
        \State Stitch the trajectory:
               $\tau_k\gets\tau_k^{\mathrm L}
               \oplus\tau_{e_k,j_k}^{\mathrm R}$
        \State Assign $A^{\mathrm L}(e_k)$ to
               $\tau_k^{\mathrm L}$ and
               $A^{\mathrm R}(e_k,j_k)$ to
               $\tau_{e_k,j_k}^{\mathrm R}$
    \EndFor
    \State \textit{// Update the policy and curriculum}
    \State Update $\theta$ by maximizing
           $\mathcal J(\theta)$ in Eq.~\ref{eq:surrogate}
    \If{$q=\mathrm{reverse}$ and
         $\widehat p_{\pi_\theta}(c_{r_i}^*)\geq\eta$}
        \If{the original starts pass the threshold
             for most training tasks}
            \State Switch to forward: $q\gets\mathrm{forward}$; $i\gets0$
        \Else
            \State Move the start backward: $i\gets i+1$
        \EndIf
    \ElsIf{$q=\mathrm{forward}$ and the schedule advances}
        \State Increase the prefix horizon: $i\gets i+1$
    \EndIf
\EndFor
\end{algorithmic}
\end{algorithm}

\section{Implementation Details}
\label{app:impl}

\subsection{Details of Training}
\paragraph{Hyperparameters for ALFWorld.}
The actor is optimized for one epoch per update using Adam with a constant learning rate of $10^{-6}$ and gradient clipping at $1.0$. Training rollouts use temperature $1.0$ and top-$p$ $0.95$. The terminal reward is binary and is scaled by 10 for training.  The reverse curriculum stride $d$ and forward curriculum stride $\Delta\ell$ are both 3, and the reliability threshold $\eta$ is $0.6$.

\begin{table}[H]
\centering
\caption{Hyperparameters of ALFWorld and WebShop.}
\label{tab:app-hp}
\small
\begin{tabular}{lcc}
\toprule
Config & ALFWorld & WebShop \\
\midrule
\texttt{train\_batch\_size}                & 32        & 32        \\
\texttt{n\_samples\_per\_prompt}           & 8         & 8         \\
\texttt{ppo\_mini\_batch\_size}            & 16        & 16        \\
\texttt{ppo\_micro\_batch\_size\_per\_gpu} & 2         & 1         \\
\texttt{ppo\_max\_token\_len\_per\_gpu}    & 12288     & 15360     \\
\texttt{log\_prob\_max\_token\_len\_per\_gpu} & 12288  & 15360     \\
\texttt{log\_prob\_micro\_batch\_size\_per\_gpu} & 2    & 1         \\
\texttt{use\_dynamic\_bsz}                 & False     & False     \\
\texttt{ulysses\_sequence\_parallel\_size} & 1         & 1         \\
\texttt{tensor\_model\_parallel\_size}     & 1         & 1         \\
\midrule
\texttt{temperature} (rollout)            & 1.0       & 1.0       \\
\texttt{top\_p} (rollout)                  & 0.95      & 0.95      \\
\texttt{temperature} (val)                 & 0         & 0         \\
\texttt{top\_p} (val)                      & 1.0       & 1.0       \\
\midrule
\texttt{actor\_learning\_rate}             & $10^{-6}$ & $10^{-6}$ \\
$\beta_1, \beta_2$ / weight\_decay         & $0.9, 0.999$ / $0.01$ & $0.9, 0.999$ / $0.01$ \\
\texttt{grad\_clip}                        & 1.0       & 1.0       \\
\texttt{max\_epochs}                       & 1         & 1         \\
\texttt{lr\_scheduler}                     & constant  & constant  \\
\midrule
\texttt{max\_prompt\_length}               & 2048      & 2048      \\
\texttt{max\_response\_length}             & 10240     & 13312     \\
\texttt{max\_model\_len}                   & 12288     & 15360     \\
\texttt{multi\_turn\_max\_turns}           & 50        & 15        \\
\texttt{action\_budget}                    & 120       & 80        \\
\midrule
\texttt{clip\_ratio}                       & 0.2       & 0.2       \\
\texttt{clip\_ratio\_c} (dual-clip)        & 3.0       & 3.0       \\
\texttt{use\_kl\_loss} / \texttt{kl\_coef} & False / 0 & False / 0 \\
\texttt{use\_kl\_in\_reward}               & False     & False     \\
\texttt{entropy\_coeff}                    & 0         & 0         \\
\texttt{loss\_agg\_mode}                   & \multicolumn{2}{c}{seq-mean-token-mean} \\
\texttt{norm\_adv\_by\_std\_in\_grpo}      & True      & True      \\
\midrule
Reverse curriculum stride ($d$)            & 3         & 3           \\
Forward curriculum stride ($\Delta\ell$)   & 3         & 3           \\
Reliability threshold ($\eta$)             & 0.6       & 0.78             \\
\midrule
Reward type                                & Binary    & Binary    \\
Reward scale                               & 10        & 10        \\
\midrule

\end{tabular}
\end{table}

\paragraph{Hyperparameters for WebShop.}
WebShop shares ALFWorld's training batch size of 32, eight samples per prompt, learning rate, rollout sampling settings, and terminal reward scale. Its maximum prompt and response lengths are 2{,}048 and 13{,}312 tokens, respectively. Both curriculum strides are 3, and the reliability threshold $\eta$ is $0.78$. Table~\ref{tab:app-hp} lists the full configuration.

\paragraph{Computing Details.}
For ALFWorld and WebShop, both Qwen2.5-1.5B-Instruct and Qwen2.5-7B-Instruct models are trained on 4$\times$NVIDIA H100 GPUs. ALFWorld is trained for 100 iterations and is evaluated on the seen task categories and instance variants, containing 140 tasks. WebShop is trained for 150 iterations and is evaluated on the test split of 500 tasks. For both environments, we sample 128 tasks at random per run and average results over 3 matched random seeds.

\subsection{Prompts}
\label{app:prompts}
\begin{figure}[H]
\centering
\includegraphics[
  width=0.59\linewidth,
  height=0.52\textheight,
  keepaspectratio
]{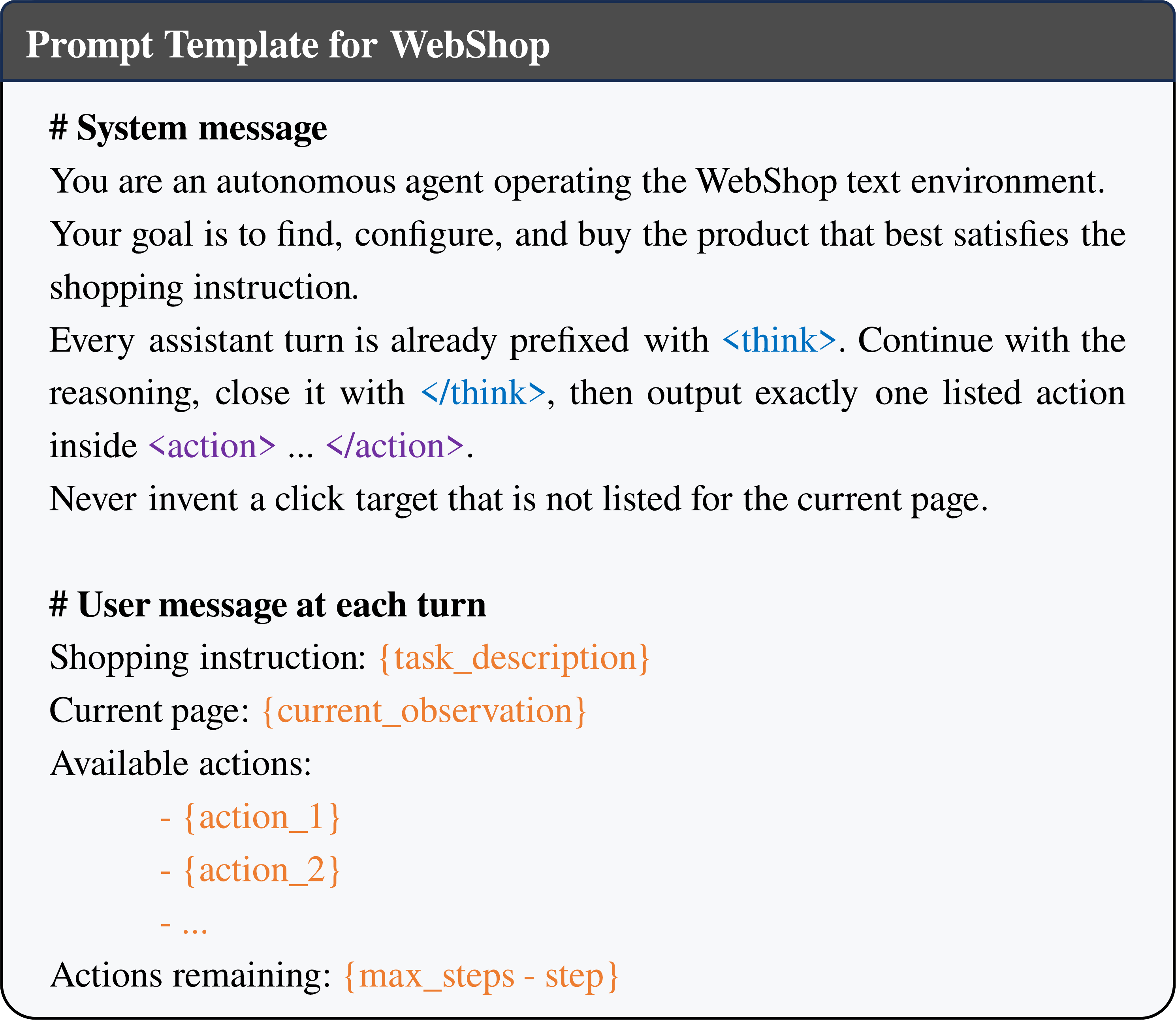}
\caption{Prompt template used for the WebShop agent.}
\label{fig:webshop-prompt}
\end{figure}

\begin{figure}[H]
\centering
\includegraphics[
  width=0.59\linewidth,
  height=0.52\textheight,
  keepaspectratio
]{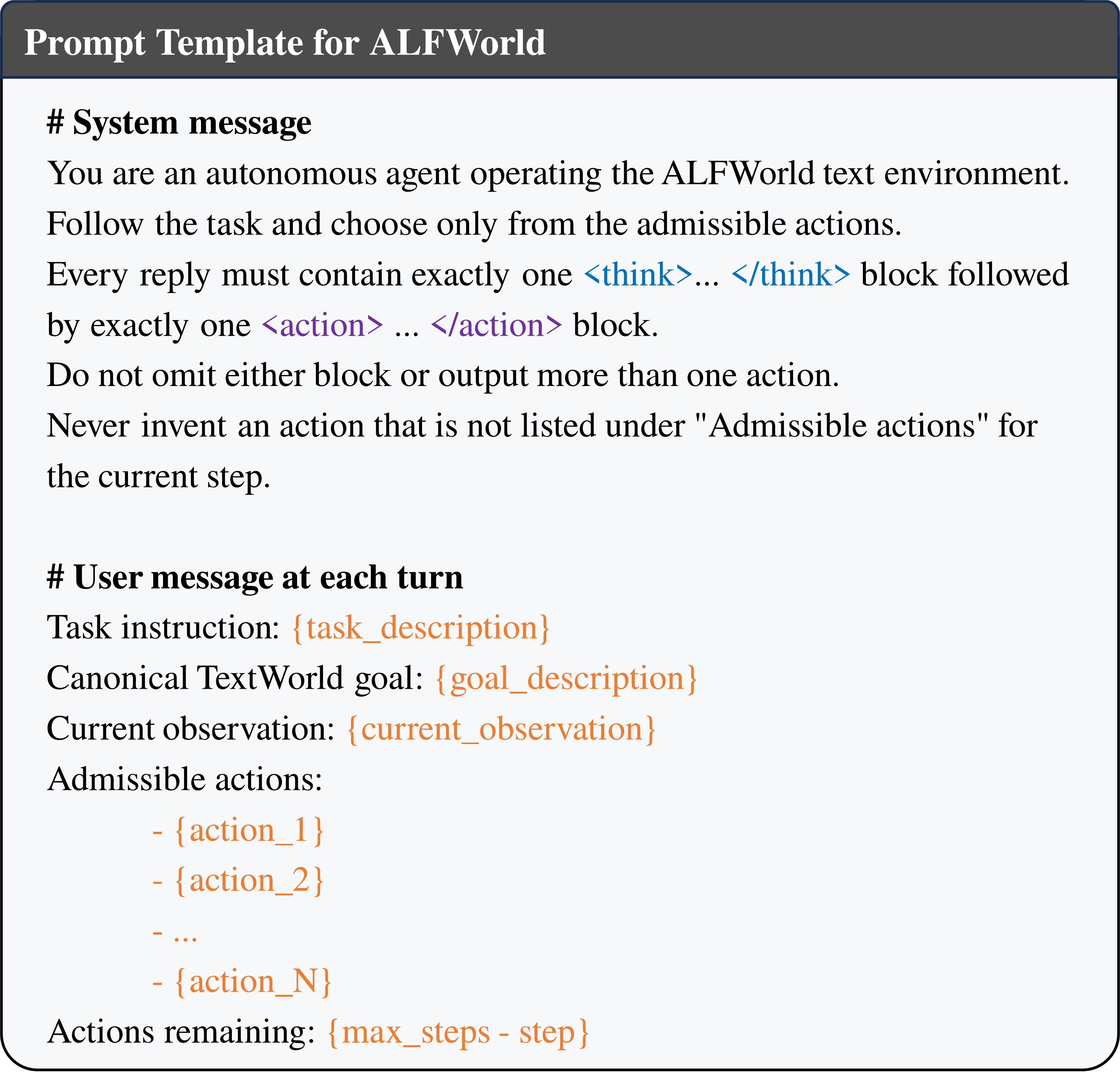}
\caption{Prompt template used for the ALFWorld agent.}
\label{fig:alfworld-prompt}
\end{figure}

Figures~\ref{fig:webshop-prompt} and~\ref{fig:alfworld-prompt}
show the prompt templates for WebShop and ALFWorld. At the start of an episode, the system message establishes the agent's task and output format. The \textcolor{promptSlot}{\texttt{\{task\_description\}}} and \textcolor{promptSlot}{\texttt{\{current\_observation\}}} denote information supplied by the environment. WebShop provides the shopping instruction and current page, while ALFWorld also provides the canonical TextWorld goal. Each environment message supplies the actions currently available and the remaining action count.

The agent then interacts within a continuing conversation. After each action, the resulting observation is added to the interaction history and the agent generates its next response from the updated context. The prompt therefore supports a structured trajectory of observations and actions rather than a separate, complete prompt for each decision. In both environments, the \textcolor{promptThink}{\texttt{<think>...</think>}} block contains the agent's reasoning, followed by exactly one listed action in the \textcolor{promptAction}{\texttt{<action>...</action>}} block. WebShop supplies the opening \texttt{<think>} tag as an assistant prefix, whereas ALFWorld requires the model to generate both complete blocks.

\section{Case Study}
\label{app:case-study}
\subsection{WebShop}
\begin{figure}[H]
    \centering
    \includegraphics[width=0.59\linewidth,height=0.52\textheight,keepaspectratio]{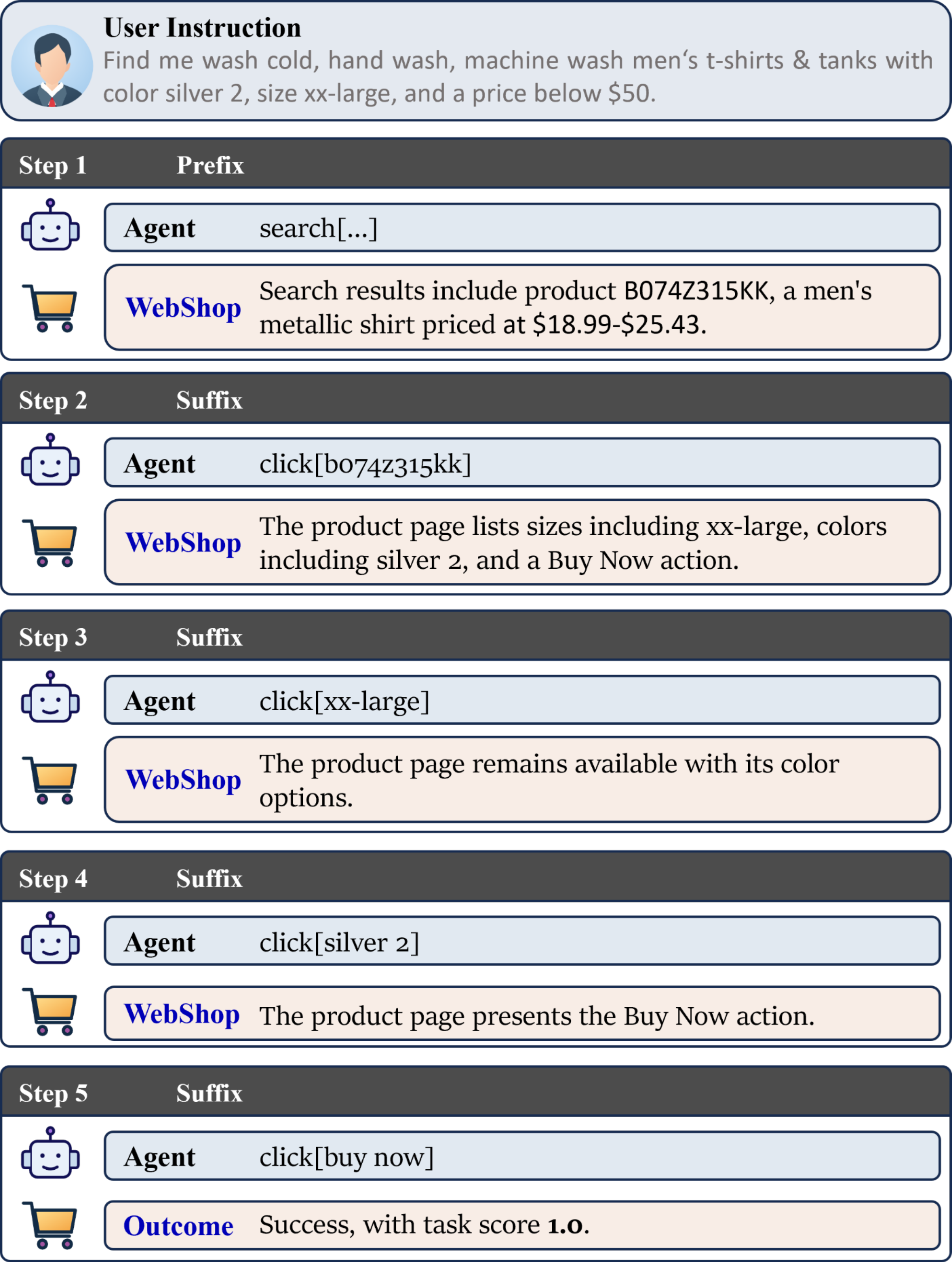}
    \caption{Illustration of a WebShop shopping task through multi-turn interaction.}
    \label{fig:webshop-case}
\end{figure}

Figure~\ref{fig:webshop-case} traces a WebShop shopping task from search to purchase. The prefix issues a search based on the user's constraints and reaches a results page with a candidate product. The suffix opens its product page, selects the requested size and color, and completes the purchase. The example shows how a search action determines the context from which the remaining decisions are made.

\subsection{ALFWorld}
\begin{figure}[H]
    \centering
    \includegraphics[width=0.66\linewidth,height=0.59\textheight,keepaspectratio]{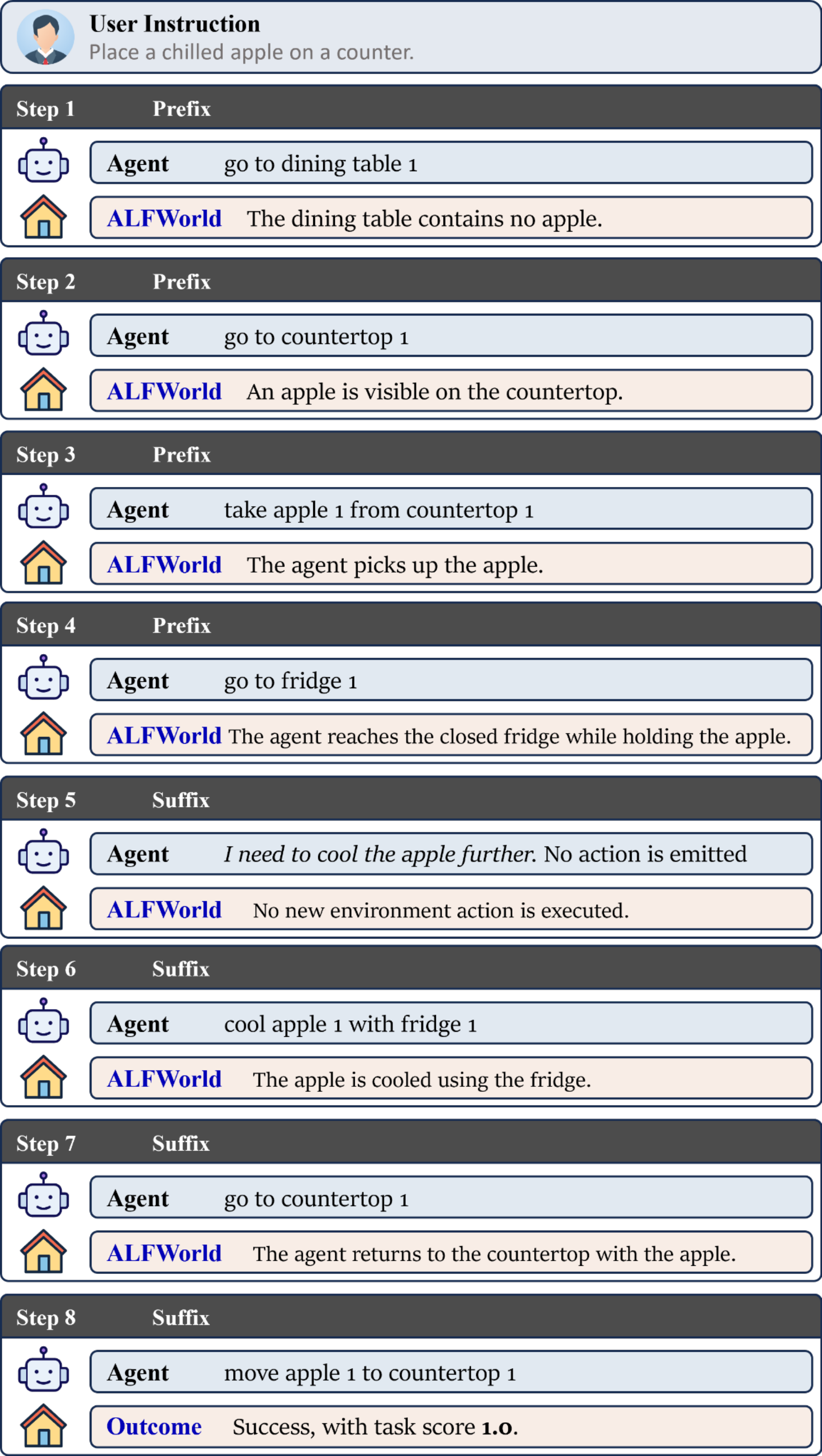}
    \caption{Illustration of an ALFWorld household task through multi-turn interaction.}
    \label{fig:alfworld-case}
\end{figure}

Figure~\ref{fig:alfworld-case} follows an agent tasked with placing a chilled apple on a counter. In the prefix, it locates and picks up the apple before reaching the fridge. The suffix cools the apple and returns it to the countertop. The recorded suffix also includes one turn in which the agent produces reasoning but takes no environment action.

\end{document}